\documentclass[conference]{IEEEtran}
\IEEEoverridecommandlockouts
\usepackage{multirow} 
\usepackage{enumitem, array} 
\usepackage{booktabs} 
\usepackage{cite}
\usepackage{amsmath,amssymb,amsfonts}
\usepackage{graphicx}
\usepackage{placeins}
\usepackage{xcolor}
\usepackage{algpseudocode}
\usepackage{subcaption}
\graphicspath{{Images/}} 
\DeclareGraphicsExtensions{.pdf,.jpg,.jpeg,.png} 
\usepackage{textcomp}
\def\BibTeX{{\rm B\kern-.05em{\sc i\kern-.025em b}\kern-.08em
    T\kern-.1667em\lower.7ex\hbox{E}\kern-.125emX}}  
\usepackage[utf8]{inputenc}
\usepackage{xcolor}

\usepackage{mathtools}
\usepackage{algpseudocode}
\usepackage{listings}
\usepackage{amsmath}
\usepackage{subcaption}
\usepackage{algorithm}
\usepackage[numbers]{natbib}
\usepackage{makecell}
\usepackage{url}
\usepackage{amsmath}

\begin{document}

\title{
A Deep U-Net Framework for Flood Hazard Mapping Using Hydraulic Simulations of the Wupper Catchment 
}

\author{\IEEEauthorblockN{
         Christian Lammers \IEEEauthorrefmark{2},
         Fernando Arévalo \IEEEauthorrefmark{1},
         Leonie Märker-Neuhaus \IEEEauthorrefmark{1},\\
         Daniel Heinenberg \IEEEauthorrefmark{1},
         Christian Förster \IEEEauthorrefmark{1},
         Karl-Heinz Spies \IEEEauthorrefmark{1}
        }
   	\IEEEauthorblockA{\IEEEauthorrefmark{1}Wupper Association (Wupperverband) \\Wuppertal, Germany\\
		Email: \{fan\}@wupperverband.de }
    \IEEEauthorblockA{\IEEEauthorrefmark{2} Institute for Computing and Information Sciences\\ Radboud University Nijmegen, Netherlands\\
        Email: christian.lammers@pm.me}        
	}

\maketitle

\begin{abstract}
The increasing frequency and severity of global flood events highlights the need for the development of rapid and reliable flood prediction tools. This process traditionally relies on computationally expensive hydraulic simulations. This research presents a prediction tool by developing a deep-learning-based surrogate model to accurately and efficiently predict the maximum water level across a grid. This was achieved by conducting a series of experiments to optimize a U-Net architecture, patch generation, and data handling for approximating a hydraulic model. 
This research demonstrates that a deep learning surrogate model can serve as a computationally efficient alternative to traditional hydraulic simulations. 
The framework was tested using hydraulic simulations of the Wupper catchment in the North-Rhein Westphalia region (Germany), obtaining comparable results.
\end{abstract}

\section{Introduction}

Over the last decades, flooding has been a growing global and local problem due to the increased land sealing and climate change, which is leading to more intense precipitation in some regions 
\citep{seneviratne2021weather,dang2024application}. For instance, the 2021 flood in Germany, especially in North Rhine-Westphalia (NRW) was considered a catastrophic century flood caused by an immense rainfall in a short duration. 
This single event claimed 49 lives and caused over 13 billion EUR in damages in NRW alone, highlighting its devastating societal and economic impact \citep{ThiekenBubeck2023}. Two of the study areas used throughout this research, Beyenburg (BEY) and Kolfurth (KOL), were among the most severely impacted regions \citep{FloodWuppertal2025}, making them critical case studies for developing and validating next-generation technologies. However, traditional physics-based models are too slow for this real-time warning due to their high computational demand \citep{BurrichterHofmann2023, FathiLiu2025}. This extreme flooding event thus highlights the urgent need for faster, more efficient alternatives. In response to these severe events and the urgent need for real-time warnings, the "Bergisches Hochwassermeldesystem 4.0" (HWS 4.0) was launched as an initiative by the Ministry of Economy in NRW and led by key regional stakeholders \citep{KINRW2025, BHWS2025}. 
While the broader HWS 4.0 project aims to forecast river discharge, an effective warning system requires mapping these discharge forecasts onto spatial flood inundation maps. 
This research provides a critical complement by developing a deep learning-based surrogate model that takes discharge as input and rapidly predicts the resulting flooding extent. 
The ultimate goal is a generalizable framework that can be deployed across various topographies.






Despite the accuracy of physics-based numerical models (PBNM) in the prediction of flood extent and depth, these PBNM are computationally expensive and time-consuming \citep{FraehrWang2024}. These limitations render those models inapplicable to emergency situations, which require rapid decision-making. Surrogate models could be an alternative to computationally intensive simulations of PBNM by approximating the relationship between, e.g., hydraulic and topographical features and water level. The primary motivation is that surrogate models can provide faster forecasts and are thus better suited for real-time flood prediction \citep{dang2024application}. 

Surrogate models, often data-driven, have the potential not only to approximate these physics-based models but also to significantly reduce the time required to simulate flooding extent \citep{dang2024application,zahura2020training}. 
Recent literature has shown that deep-learning-based surrogate models can accurately approximate the extent of flooding at locations where they have been trained \citep{bentivoglio2022deep, ChoiWoo2025}. However, deploying these models is challenging because they might overfit to the domain's unique topographical and hydraulic characteristics. 

Surrogate models for hydraulic modeling have been developed using data-driven methods that employ machine learning (ML) algorithms such as random forests, Gaussian processes, least-squares support vector machine regression, and neural networks \citep{zahura2020training, dang2024application, BermudezCea2019}.
One of the main advantages of ML is its ability to construct an input-output relationship without relying on physical reinforcement \citep{herath2021genetic}. Additionally, ML models require less extensive development and calibration work than physics-based models.
\citet{zahura2020training} developed a surrogate model based on a random forest (RF) algorithm. The authors trained the RF on various topographical and environmental features to forecast hourly water depths that were initially simulated by a 1-D/2-D physics-based model. Their study demonstrated that the RF model could provide accurate water level forecasts while achieving a runtime 3000 times faster than the physics-based model. These results highlight the significant computational advantage of using machine learning techniques for surrogate modeling, which naturally raises the question of how RF compares to other available algorithms for this task. 
\citet{dang2024application} conducted a comparative study of nine different classical machine learning models for predicting flood depth. While their analysis confirmed that models such as Gaussian Processes (GP), Random Forests (RF), and Neural Networks (NN) achieved the highest performance, the authors' approach had a key limitation. It relied on simplified inputs such as water and tidal levels, ignoring critical spatial information available from sources like Digital Elevation Models (DEMs). The authors concluded that prediction performance could be significantly improved by employing more advanced deep learning architectures, specifically Convolutional Neural Networks (CNNs).
Alternatively, \citet{herath2021genetic} explored hybrid approaches that bridge the gap between purely physics-based and data-driven models. Their framework uses Genetic Programming (GP) to incorporate existing hydrological knowledge directly into the machine learning algorithm. This method produces models that are not only accurate for tasks such as rainfall-runoff forecasting but also more interpretable to domain experts. However, this approach has notable limitations for real-time applications. By incorporating complex hydrological concepts, the focus shifts towards interpretability, often at the expense of raw computational speed, which is needed for rapid emergency forecasting.

The limitations identified by \citet{dang2024application} highlight the need to explore more sophisticated deep learning architectures such as convolutional neural networks (CNN), temporal graph convolutional network (T-GCN), and transformers (e.g., the combination of CNN and LSTM) \citep{kabir2020deep, BurrichterHofmann2023}.
\citet{kabir2020deep} developed a 1D-CNN for rapid fluvial flood prediction and benchmarked it against a classical Support Vector Regression (SVR) model. Their results demonstrated the CNN's superior ability to capture the spatial extent of inundated areas compared to the interpolated SVR maps. This was further supported by the CNN reducing the 99th-percentile error to 0.2 m, compared to 1.6m for the SVR. However, the study's scope was limited to demonstrating the model's ability to generalize across different rainfall events within a single location. This leaves the critical question of generalization across different spatial flood plains unresolved. 
Likewise, \citet{guo2022data} trained a CNN on multiple catchments to test its performance on unseen terrains. Their study investigated how to handle inputs of varying sizes, comparing a resizing-based method with a patch-based method. They found the latter to be more effective in terms of prediction accuracy. While the study successfully demonstrated that their model could generalize spatially, it also had a significant limitation: it was trained and tested on a single rainfall event. This leaves open the question of how such a model would perform under different hydrological conditions. 

To this end, U-Net has shown particular promise in generalizing across different catchments \citep{hosseiny2021deep, kabir2023deep}. \citet{hosseiny2021deep}, for instance, employed a U-Net for water depth estimation and demonstrated a 29\% improvement in accuracy compared to a conventional ANN trained on the same data. This result helped establish the U-Net as a powerful and highly effective architecture for this task. However, the study identified a critical limitation: data availability, particularly the difficulty of acquiring sufficient training data when applying the model to a new study area. This highlights the core problem of generalization, a central challenge for the practical application of such deep learning models. 
\citet{kabir2023deep} demonstrated that a U-Net trained on multiple catchments could generalize to unseen topographies and storm events. For this, the authors trained their model on patches of size $1024\times1024$. However, their approach revealed a significant trade-off. While the model showed generalization potential, it came at the cost of an immense training time of 20 days and a tendency to underestimate flood depths and extents consistently.

Though the U-Net-based architectures have shown significant results on hydraulic modeling, the data required to train the models is often extensive and typically includes larger images of the catchment. This poses a substantial challenge because the training process requires intensive computing resources and longer computation times. To this end, different strategies have been proposed, such as resizing and dividing images into patches \citep{guo2022data, cache2024enhancing}. For instance, \citet{guo2022data} compared a patching vs. resizing strategy for handling large areas and found that the patching strategy is the better option, particularly for flood prediction. The approach of breaking large images into more digestible chunks has also been confirmed by \citet{ronneberger_u-net_2015} and \citet{masci2015geodesic}. 

To address the aforementioned gaps, this article proposes a deep U-Net framework for flood hazard mapping using hydraulic simulations. In addition, it provides a methodology to test the model's robustness by applying different inference patch strategies. Moreover, it presents an ablation study of patch parameters and target normalization to determine their impact on the model performance. The framework is tested using simulations of selected locations of the Wupper catchment.
The framework methodology is detailed in terms of data preparation, training process, and inference strategies, thereby facilitating practical implementation in similar research projects.

\section{Methodology}
This section describes the study site, data generation, application of the U-Net for flood hazard mapping, model evaluation using different inference strategies, and evaluation metrics.  







\subsection{Application study site}
\begin{figure*}[!htbp]
	\centering
	\includegraphics[width=0.9\textwidth,keepaspectratio]{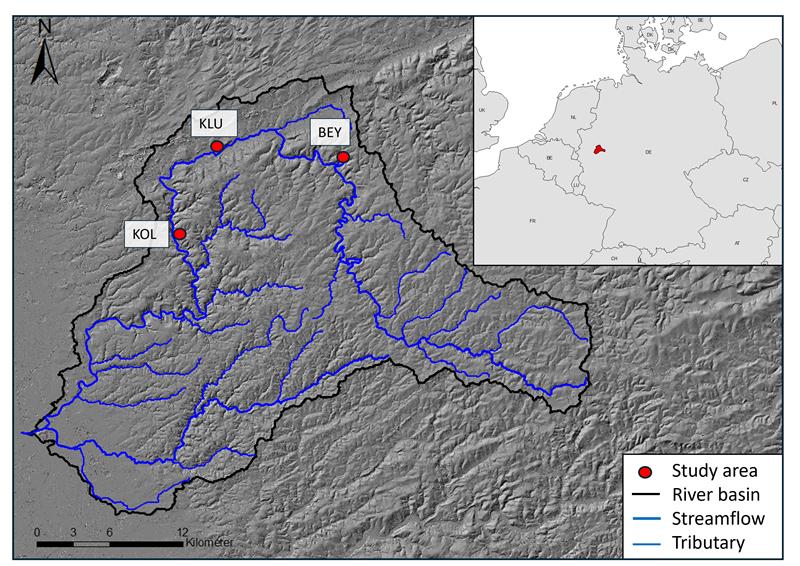}
	\caption{Study areas: Beyenburg (BEY), Kluserbrücke (KLU), and Kohlfurth (KOL)}\label{Fig__use_case}
\end{figure*}

The study area is located in the catchment area of the Wupper, a river that rises in the Oberbergischer Kreis district and flows into the Rhine near Leverkusen after a course of 117 km.
In this area, we chose three distinct locations: Beyenburg (BEY), Kluserbrücke (KLU), and Kolfurth (KOL). These study areas can be seen on the map in Fig. \ref{Fig__use_case}, their geographical characteristics in Table \ref{tab_study_area_chars_geo}, and hydrologic characteristics in Table \ref{tab_study_area_chars_hydro}. For this work, BEY serves as the primary domain for model development and approximation, KLU is the target domain for transfer learning, and KOL is used as final domain to evaluate broader generalization to unknown topography. All of these three locations, especially the areas around Beyenburg and Kolfurth have been hit severely by the flooding event from 2021. Thus, due to their real-world importance in the context of flooding and their unique topographical challenges, these locations are ideal candidates for this study. 

\begin{table*}[!htbp]
    \centering
    \caption{Geographical characteristics of the three study areas.}
    \label{tab_study_area_chars_geo}    
    \begin{tabular}{lccccc} 
        \toprule
        \textbf{Location} &
        \textbf{\makecell{Area\\($km^2$)}} &
        \textbf{\makecell{Elevation\\(m NHN)}} &
        \textbf{\makecell{Image size}} &
        \textbf{\makecell{Avg. Sim.\\Time (min)}} \\
        \midrule
        Beyenburg (BEY)     & 1.73 & 183.25 -- 226.39 & $2041\times849$     & 15 \\
        Kluserbrücke (KLU) & 6.88 & 117.28 -- 194.51 & $10402\times6617$   & 20 \\
        Kohlfurth (KOL)     & 3.77 & 91.83 -- 175.91  & $3470\times10877$   & 17 \\
        \bottomrule
    \end{tabular} 
\end{table*}

\begin{table*}[!htbp]
    \centering
    \caption{Hydrological characteristics of the three study areas.}
    \label{tab_study_area_chars_hydro}  
    \begin{tabular}{lcccc} 
        \toprule
        \textbf{Location} & \textbf{Bankfull} ($m^3/s$) & \textbf{$HQ_{100}$ ($m^3/s$)} & \textbf{$HQ_{extreme}$ ($m^3/s$)} \\
        \midrule
        Beyenburg (BEY)     & 65-70   & 106  & 160 \\
        Kluser Brücke (KLU) & 190  & 172 & 258 \\
        Kohlfurth (KOL)     & 130  & 201 & 302 \\
        \bottomrule
    \end{tabular}
\end{table*}

\subsection{Data Generation using a Patch Strategy}

The training, validation, and testing pairs for the data-driven model consist of a three-channel input (elevation, discharge, and a binary mask) and a one-channel output indicating the maximum water level at each grid point.

A digital elevation model (DEM) was used as the primary topographical input, providing elevation data, which has also been used in similar work \citep{singha2025advancing,guo2022data}.
This DEM is publicly available \citep{geobasisNRWDGM2025}. However, since it comes only as individual tiles, they must be downloaded, post-processed, and then stitched together. Therefore, individual tiles have been downloaded and adapted to obtain the whole study area for each of the three locations. Adapting involved stitching the tiles together and then cropping to match the exact study area used to simulate the water level. It is important to note that while the DEM was stitched and cropped, not all man-made structures, such as bridges, were removed from the elevation data used to train the surrogate model. The resulting GeoTIFF file is a topographic map with a single band indicating elevation above mean sea level, at a resolution of 1x1 meters. 
The underlying coordinate system is the EPSG:25832. This raster image was used as the first channel for the U-Net.

The second channel consisted of a constant river discharge value used to generate the flooding simulation. 
The scalar was broadcast to the same shape as the elevation map, and both channels were concatenated. 
Additionally, a mask was generated to indicate invalid/no-data regions. These invalid data areas, are those areas for which the hydraulic model did not output any data. 
This binary mask, thus, indicates valid and invalid grid points and was added as a third channel to the input stack. This process is quite common in the  literature 
\citep{guo2022data,guo2021data}. 
Lastly, all input data has been normalized. This process of normalizing the data helps the model to converge faster and avoids exploding gradients. 
We have chosen for min-max scaling as this has proven itself according to recent research \citep{cache2024enhancing, FarfanMontalvo2025}. Elevation was normalized across the entire study area rather than patch-wise. Additionally, discharge values were normalized using min-max scaling. Following best practices, the statistics used to normalize the input were calculated only from the training set. This was done to prevent any leakage of the test set into the training phase. The same statistics were then applied to transform the validation and test sets, including the unseen locations.
The target used for training the data-driven model is the flood inundation map simulated by the hydraulic model. 
This simulation is in the shapefile format, a standard format for storing map data consisting of points, lines, and polygons. To use this shapefile as training data, it was first rasterized into a raster file, which was then used to train the surrogate model. This was done with the GIS software ArcGIS Pro.

\subsubsection{Patch Generation Strategy}
As shown in Table \ref{tab_study_area_chars_hydro}, the three study areas BEY, KLU, and KOL cover an area of 1.73 $km^2$, 6.88 $km^2$, and 3.77 $km^2$, which results in image sizes of $849 \times 2041$, $6617\times10402$, and $10877 \times 3470$, respectively. Processing a complete image of size $849\times2041$, as is the case for the study area BEY, is infeasible due to the considerable computational resources required and the long training times. For these reasons, we adopted a patching strategy. Similar strategies have been applied by \citet{guo2022data} and \citet{cache2024enhancing} and deemed quite efficient. 

Therefore, patches were created for each input-target pair and used to train the model. These patches were randomly sampled across the study area, and each patch contained at least one valid data point. This patch sampling follows the standard rejection sampling method. A detailed overview of this procedure is provided in Algorithm \ref{alg:patch_sampling}.
\begin{algorithm}
\caption{Valid Patch Extraction via Rejection Sampling}
\label{alg:patch_sampling}
\begin{algorithmic}[1]
\Require Dataset of full rasters $\mathcal{D}$, Global Valid Mask $\mathcal{M}$
\Require Patch Size $P$, Min Valid Pixels $\tau$
\Ensure A valid training patch $(x, y)$

\State $idx \sim \text{Uniform}(0, |\mathcal{D}|)$ \Comment{Select random image index}
\State $X_{full}, Y_{full} \gets \mathcal{D}[idx]$ \Comment{Retrieve cached full image}

\Repeat
    \State $r \sim \text{Uniform}(0, H - P)$ \Comment{Random row}
    \State $c \sim \text{Uniform}(0, W - P)$ \Comment{Random col}
    
    \State $Pixels \gets \text{CountValidPixels}(\mathcal{M}, r, c, P)$
    
\Until{$Pixels \ge \tau$}

\State $x \gets X_{full}[r:r+P, c:c+P]$
\State $y \gets Y_{full}[r:r+P, c:c+P]$
\State \Return $(x, y)$
\end{algorithmic}
\end{algorithm}

Based on the work of \citet{hosseiny2021deep}, an initial patch size of $128\times128$ was selected as a baseline for the experiments. This size is adequate for similar tasks and for the baseline architecture. Additionally, we investigated how the number of patches sampled from a single input-target pair affected training performance and training time. 

\subsubsection{Data Augmentation Strategy}

To increase the robustness and generalization of our surrogate model and to incorporate regularization, we augmented the training dataset. 
Unlike prior work, in which augmentation was primarily applied to address data scarcity, our primary motivation was to enhance the model's ability to generalize by creating a more diverse set of training examples.
The reasoning behind this approach is that training a model on augmented patches, especially rotations and horizontal and vertical flips, would make the model invariant to topographical orientation and thus focus on more robust features, ideally enhancing generalization to unseen locations. 
For this, we applied the following three data augmentation strategies to the training patches: horizontal flipping, vertical flipping, and 90°, 180°, or 270 ° rotation.
Each of these augmentation strategies was applied with a specific probability. To preserve the physical relationship between topographical features and water level, input and target patches in the training set were simultaneously augmented \citep{cache2024enhancing}. 

\subsection{A Deep U-Net Framework for Flood Hazard Mapping}
The U-Net is a CNN architecture designed for tasks that require pixel-level output, such as segmentation or flood mapping \citep{ronneberger_u-net_2015}. It works on an encoder-decoder principle, which can be thought of as a two-stage process of summarizing and then reconstructing the input image. The first stage sends an input image through a series of convolutional and max pooling layers through a so-called encoder. With each step in this encoder, the image resolution shrinks while the number of feature maps grows. At the end of the encoder, a so-called bottleneck is applied to connect the first stage to the second stage. The second stage, also referred to as the decoder, increases the resolution again and reduces the number of feature maps. The final output of the decoder is an image with the exact spatial resolution as the original input image.
Usually, the depth of such a U-Net refers to the number of these downsampling and upsampling blocks. Additionally, the width of a network refers to the number of filters in the initial layer. 
The model has a depth of 4 and a width of 16. Details of the U-Net architecture can be found in \citet{hosseiny2021deep}.








Skip connections are an essential concept of the U-Net architecture. These connections bridge the encoding and decoding path. While the encoder reduces the resolution of the data, high-resolution spatial information might be lost, which could otherwise benefit the decoder for more precise pixel-level prediction. To overcome this, the feature maps from the encoder stage are concatenated with the upsampled feature maps from the corresponding decoder stage.
The skip connections make the U-Net especially useful in the domain of flood hazard mapping, as they provide fine-grained topographic details, such as riverbanks and small structures, that are highly relevant to the task. 
The actual number of encoder and decoder blocks, referred to as the network depth, and the initial filter count, referred to as the width, have been systematically optimized through a series of preliminary experiments. 
\begin{algorithm}
\caption{Training Procedure for Water Depth Estimation}
\label{alg:training}
\begin{algorithmic}[1]
\Require Training Set $\mathcal{D}_{train}$, Validation Set $\mathcal{D}_{val}$, Configuration $C$
\Ensure Trained Model $\mathcal{M}^*$

\State Initialize Model $\mathcal{M}$ with parameters $\theta$ (U-Net)
\State Initialize Optimizer and Learning Rate Scheduler
\For{$epoch = 1$ \textbf{to} $C_{epochs}$}
    \State \textbf{Training Phase:}
    \State Set $\mathcal{M}$ to training mode
    \For{$step = 1$ \textbf{to} $Steps_{per\_epoch}$}
        \State \Comment{Construct batch using Algorithm \ref{alg:patch_sampling}}
        \State $\mathcal{B} \gets \emptyset$
        \For{$i = 1$ \textbf{to} $BatchSize$}
            \State $p_i \gets \text{SampleValidPatch}(\mathcal{D}_{train}, P, \tau)$
            \State $\mathcal{B} \gets \mathcal{B} \cup \{p_i\}$
        \EndFor
        
        \State $(X, Y, Mask) \gets \text{Stack}(\mathcal{B})$
        \State $X, Y \gets$ Normalize($X, Y$
        \State $\hat{Y} \gets \mathcal{M}(X)$ 
        \State $Loss \gets \text{MaskedRMSE}(\hat{Y}, Y, Mask)$
        \State Zero Gradients
        \State Backpropagate $Loss$
        \State Optimizer Step
    \EndFor
    \State Scheduler Step

    \State \textbf{Validation Phase:}
    \State Set $\mathcal{M}$ to evaluation mode
    \State $L_{val} \gets 0$
    \For{batch $(X, Y, Mask)$ \textbf{in} $\mathcal{D}_{val}$}
        \State $\hat{Y} \gets$ SpatialInference($\mathcal{M}, X$) 
        \State $L_{val} \gets L_{val} + \text{MaskedRMSE}(\hat{Y}, Y, Mask)$
    \EndFor
    \State $L_{val} \gets \text{Average}(L_{val})$

    \State \textbf{Check Early Stopping:}
    \If{$L_{val} < L_{best}$}
        \State $L_{best} \gets L_{val}$
        \State $\theta_{best} \gets \theta$ 
        \State $counter_{early\_stop} \gets 0$
    \Else
        \State $counter_{early\_stop} \gets counter_{early\_stop} + 1$
    \EndIf
    
    \If{$counter_{early\_stop} \ge C_{patience}$}
        \State \textbf{break}
    \EndIf
\EndFor
\State \Return $\mathcal{M}$ loaded with $\theta_{best}$
\end{algorithmic}
\end{algorithm}
This model was chosen for its ability to capture the complex spatial relationships among topography, discharge, and water level. At the same time, the model can preserve the spatial precision needed for high-resolution water-level prediction, achieved through skip connections and an encoder-decoder structure. 
The U-Net has proven to be useful for approximating hydraulic models, particularly for predicting water levels \citep{hosseiny2021deep, guo2022data, cache2024enhancing}. 
We have chosen the exact implementation of the U-Net as described by \citet{hosseiny2021deep}, which contains four encoder and decoder blocks, with the first layer containing 16 filters. 
The encoder block consists of two convolutional layers, each with a kernel of $3\times3$. After each convolutional layer, a batch normalization layer is applied, followed by a ReLU activation, introducing nonlinearity into the model. Finally, a max pooling layer with a $2\times2$ filter and a stride of 2 is applied to reduce the spatial resolution of the feature maps. A decoder block consists of an up-convolutional layer with a kernel size of $2\times2$ and a stride of 2, followed by two instances of a convolutional layer, each with a kernel size of $3\times3$, batch normalization, and ReLU activation function. The output layer consists of a single $1\times1$ convolutional layer with a single channel, producing the predicted water level at each grid point.
In addition, different measures are implemented to mitigate overfitting, such as early stopping \citep{PianforiniDazzi2024, FarfanMontalvo2025} and data augmentation \citep{DofitasKim2025, ChoiWoo2025}. Algorithm \ref{alg:training} shows a detailed sequence of the training process.


\subsection{Inference Strategy}

To evaluate the model's performance across the entire study area and to mitigate boundary artifacts, three distinct inference strategies were tested. Validation and testing images were processed as a whole rather than using random patches. The three tested strategies were:
\begin{itemize}
    \item No overlap: The image was split into adjacent, non-overlapping patches.
    \item Overlap: The image was split into patches using a stride of 50\% of the patch size. The predictions in these overlapping areas were then averaged.
    \item Center Crop: Only the central portion of each patch's prediction was used for the final map.
\end{itemize}

The non-overlapping patch strategy is shown in Algorithm \ref{alg:non_overlap}. For this strategy, prediction patches are placed back into the original image. Each patch window is then slid across the input area by precisely the same size as the patch. By doing this, prediction patches are not overlapping, and every single pixel receives one prediction value. This simple tiling approach serves as a common baseline in patch-based segmentation \citep{HuangReichman2018}
\begin{algorithm}
\caption{Inference Strategy: Non-Overlapping Patches}
\label{alg:non_overlap}
\begin{algorithmic}[1]
\Require Image $I$ of size $(H, W)$, Model $\mathcal{M}$, Patch Size $P$
\Ensure Prediction map $O$

\State Calculate padding to make $(H, W)$ multiples of $P$
\State $I_{pad} \gets \text{ReflectPad}(I)$
\State Initialize empty result grid $O_{pad}$

\For{$y = 0$ \textbf{to} $H_{pad}$ \textbf{step} $P$}
    \For{$x = 0$ \textbf{to} $W_{pad}$ \textbf{step} $P$}
        \State $patch \gets I_{pad}[y : y+P, x : x+P]$
        \State $pred \gets \mathcal{M}(patch)$
        \State $O_{pad}[y : y+P, x : x+P] \gets pred$ \Comment{Direct placement}
    \EndFor
\EndFor

\State $O \gets \text{Crop}(O_{pad}, H, W)$
\State \Return $O$
\end{algorithmic}
\end{algorithm}

The overlapping patch strategy is shown in Algorithm \ref{alg:sliding_window}. Instead of sliding the patch window by the same size as the patch, in this strategy, the patch window is only slid by half of the patch size. This results in some pixels receiving two predictions. The final prediction is the average of these overlapping predictions, a common approach to mitigate boundary artifacts in patch-based semantic segmentation of large remote sensing images \citep{HuangReichman2018}.
\begin{algorithm}
\caption{Inference Strategy: Sliding Window (Overlapping)}
\label{alg:sliding_window}
\begin{algorithmic}[1]
\Require Image $I$, Model $\mathcal{M}$, Patch Size $P$, Stride $S$ ($S < P$)
\Ensure Prediction map $O$

\State Calculate padding to make $(H, W)$ multiples of $S$
\State $I_{pad} \gets \text{ReflectPad}(I)$
\State Initialize Accumulator $A$ and Counter $C$ with zeros

\For{$y = 0$ \textbf{to} $H_{pad}-P$ \textbf{step} $S$}
    \For{$x = 0$ \textbf{to} $W_{pad}-P$ \textbf{step} $S$}
        \State $patch \gets I_{pad}[y : y+P, x : x+P]$
        \State $pred \gets \mathcal{M}(patch)$
        
        \State \Comment{Accum. pred. in overlapp. regions}
        \State $A[y : y+P, x : x+P] \gets A[y : y+P, x : x+P] + pred$
        \State $C[y : y+P, x : x+P] \gets C[y : y+P, x : x+P] + 1$
    \EndFor
\EndFor

\State $O_{pad} \gets A / C$ \Comment{Average the overlaps}
\State $O \gets \text{Crop}(O_{pad}, H, W)$
\State \Return $O$
\end{algorithmic}
\end{algorithm}

Unlike the other two strategies, the center crop strategy, as shown in Algorithm \ref{alg:center_crop}, considers only the center predictions of each patch. These center predictions are then placed into the original image, discarding the outer predictions of each patch. This idea is closely related to the overlap–tile strategy used in U‑Net for seamless segmentation of large images \citep{ronneberger_u-net_2015}.
\begin{algorithm}
\caption{Inference Strategy: Center Cropping}
\label{alg:center_crop}
\begin{algorithmic}[1]
\Require Image $I$, Model $\mathcal{M}$, Patch Size $P_{total}$, Center Size $P_{center}$
\Ensure Prediction map $O$

\State $pad_{context} \gets (P_{total} - P_{center}) / 2$
\State Pad $I$ so dimensions are multiples of $P_{center}$
\State $I_{pad} \gets \text{ReflectPad}(I, pad_{context})$ \Comment{Add extra context padding}

\For{$y = 0$ \textbf{to} $H_{pad}$ \textbf{step} $P_{center}$}
    \For{$x = 0$ \textbf{to} $W_{pad}$ \textbf{step} $P_{center}$}
    
        \State \Comment{Extract larger patch including context borders}
        \State $patch \gets I_{pad}[y : y+P_{total}, x : x+P_{total}]$
        \State $pred_{full} \gets \mathcal{M}(patch)$
        
        \State \Comment{Discard borders, keep only the valid center}
        \State $pred_{center} \gets pred_{full}[pad_{context} : -pad_{context}, pad_{context} : -pad_{context}]$
        
        \State $O[y : y+P_{center}, x : x+P_{center}] \gets pred_{center}$
    \EndFor
\EndFor

\State $O \gets \text{Crop}(O, H, W)$
\State \Return $O$
\end{algorithmic}
\end{algorithm}

\subsection{Evaluation Metrics}

The root mean squared error (RMSE) was selected because it provides a clear interpretation in the same units as the target data and additionally penalizes larger errors more severely than, e.g., the MSE loss. This penalization of significant errors is preferable for this study, as the goal is not only to approximate the hydraulic model but also to provide a framework for real-world use. Therefore, penalizing significant errors is particularly desirable for flood modeling, where significant underpredictions of water levels could lead to failed risk assessments and inadequate safety measures. Several other authors in the same field have used RMSE as a test metric \citep{kabir2020deep,guo2022data,bentivoglio2022deep, FathiLiu2025}. The RMSE is calculated using:

\begin{equation}
RMSE = \sqrt{\frac{1}{n}\sum_{i=1}^{n}(y_{pred_i} - y_{true_i})^2}
\label{eq:rmse}
\end{equation}
In addition to the RMSE, we also report the Nash-Sutcliffe efficiency (NSE) score to provide a comprehensive evaluation of the model's predictive power \citep{nash1970river}. It is commonly applied as a calibration metric when calibrating hydraulic models \citep{XuDeVos2025, LiDing2026}:
\begin{equation}
    \label{eq:nse}
    \text{NSE} = 1 - \frac{\sum_{i=1}^{n}(y_{\text{pred}_i} - y_{\text{true}_i})^2}{\sum_{i=1}^{n}(y_{\text{true}_i} - \overline{y}_{\text{true}})^2}
\end{equation}

The NSE score quantifies the model's predictive skill relative to the mean of the observed data. While the observed data typically come from field measurements, in this surrogate modeling study, the 'observations' refer to the ground-truth simulations generated by the hydraulic model. An NSE score of 1 indicates a perfect match, a score of 0 indicates the model is no better than the mean of the observations, and a negative score indicates the model performs worse than using the mean of the observations.

\section{Results}

This section presents the results of data generation and preparation, highlighting the patch-generation and data-augmentation strategies. In addition, the results of the U-Net-based model approximation are presented, focusing on patch overlap during inference, along with a qualitative assessment. An ablation study describes the impact of patch parameters and target normalization during model training.   


\subsection{Data Generation and Preparation}
To generate the training data for this study, the 2D simulation software HydroAS was used \citep{HydroAS2025}.
This model numerically solves the 2D shallow-water equations, which are fundamental physical equations describing the flow of fluids. 
In this hydraulic model, the riverbank topography is represented by a digital terrain model (DTM), and the river channel by integrated survey data of the watercourse. Other parameters, such as roughness, are also taken into account. 
A given discharge serves as input for the model. For the convenience of this study, a constant discharge, i.e., a steady state, was used for the calculations.
Based on this data, the model computes the water level for each grid cell. The results are simulated flood inundation maps with water levels for each given discharge.
For each study area, a series of simulations was conducted across an array of discharge values. These discharge values for each location are shown in Table \ref{tab:data_splits}. 
Lastly, the raw simulation outputs were pre-processed. Grid points with no simulation data, either because the water would not reach this place or because they were outside the study area, were filled with "Not a Number" (NaN) values. 

\begin{table*}[!htpb]
    \centering
    \caption[Overview of data splits]{Data splits showing discharge values ($m^3/s$) for each study area.}
    \label{tab:data_splits}
    \begin{tabular}{l p{3.5cm} p{2cm} p{3.5cm}} 
        \toprule
        \textbf{Location} & \textbf{Training} & \textbf{Validation} & \textbf{Test} \\
        \midrule
        Beyenburg (BEY) & 5, 20, 50, 80, 110, 140, 170, 200, 230, 260, 275, 290, 320, 335, 350, 365, 380, 395 & 35, 95, 155, 215 & 65, 125, 185, 245, 305 \\
        \\ 
        Kluser Brücke (KLU) & 5, 35, 65, 95, 125\, 155, 185, 215, 245, 275, 305, 335 365, 395, 410 & 50, 110, 170, 230 & 20, 80, 140, 200, 260 \\
        \\ 
        Kohlfurth (KOL) & 5, 20, 35, 80, 95, 140, 155, 200, 215, 260, 275, 290, 305, 320, 335, 350, 365, 380, 395, 410, 425, 440, 455, 470, 485, 500 & 65, 125, 185, 245 & 50, 110, 170, 230 \\
        \bottomrule
    \end{tabular}
    \par 
\end{table*}

\subsubsection{Patch Generation Strategy}
\label{subsec:patch_generation}

As shown in Table \ref{tab_study_area_chars_geo}, the three study areas BEY, KLU, and KOL cover an area of 1.73 $km^2$, 6.88 $km^2$, and 3.77 $km^2$, which results in image sizes of $849\times2041$, $6617\times10402$, and $10877\times3470$, respectively. Processing a complete image of size $849\times2041$, as is the case for the study area BEY, is unfeasible due to the considerable computational resources required and the long training times. For these reasons, we adopted a patching strategy. Similar strategies have been applied by \citet{guo2022data} and \citet{cache2024enhancing} and deemed quite efficient. 
Therefore, patches were generated for each input-target pair and used to train the model. These patches were randomly sampled across the study area, and each patch contained at least one valid data point. Based on the work of \citet{hosseiny2021deep}, an initial patch size of $128\times128$ was selected as a baseline for the experiments. This size is adequate for similar tasks and for the baseline architecture. Additionally, we investigated how the number of patches sampled from a single input-target pair affected training performance and training time.

\subsubsection{Data Augmentation Strategy} 
\label{subsec:augmentation}

To increase the robustness and generalization of our surrogate model and to incorporate regularization, we augmented the training dataset. 
Unlike prior work, in which augmentation was primarily applied to address data scarcity, our primary motivation was to enhance the model's ability to generalize by creating a more diverse set of training examples \citep{DofitasKim2025, ChoiWoo2025}. 
The reasoning behind this approach is that training a model on augmented patches, especially rotations and horizontal and vertical flips, would make the model invariant to topographical orientation and thus focus on more robust features, ideally enhancing generalization to unseen locations. 

For this, we applied the following three data augmentation strategies to the training patches: 
\begin{itemize}
    \item Flipping patches horizontally
    \item Flipping patches vertically
    \item Rotating patches by 90°, 180° or 270°
\end{itemize}

Each of these augmentation strategies was applied with a specific probability. To preserve the physical relationship between topographical features and water level, input and target patches in the training set were simultaneously augmented \citep{cache2024enhancing}. 

\subsection{Experiment settings}

All models presented in this research have been trained with the same framework. The programming language used was Python 3.10.12, along with additional essential packages such as PyTorch, NumPy, Optuna, and Scikit-Learn \citep{AnselYang2024, HarrisMillman2020, AkibaYanase2019, PedregosaVaroquaux2011}.
For training, we selected the Adam optimizer \citep{kingma_adam_2017}, the ReduceLROnPlateau scheduler, and the RMSE loss function. The specific formula for this loss function can be seen in Equation \ref{eq:rmse}. Furthermore, the exact learning rate was determined through an Optuna-based search, and all other key hyperparameters are detailed in Table \ref{tab:hyperparameters}.

\begin{table}[h!]
    \centering
    \caption[Overview of training hyperparameters]{Key training hyperparameters for the model approximation experiments.}
    \label{tab:hyperparameters}
    \begin{tabular}{ll}
        \toprule
        \textbf{Hyperparameter} & \textbf{Value} \\
        \midrule
        Optimizer               & Adam \\
        Learning Rate           & 0.0000427 \\
        Batch Size              & 32 \\
        Max Epochs              & 750 \\
        Early Stopping          & Patience of 75 epochs \\
        Scheduler               & ReduceLROnPlateau \\
        \quad - Factor          & 0.1  \\
        \quad - Patience        & 10 \\
        \bottomrule
    \end{tabular}
\end{table}

\subsection{Model Approximation using U-Net}

Following the optimization of data handling, the focus shifted to the model's architecture. This section details two experiments conducted to optimize the U-Net architecture by first varying the model's depth, or encoder \& decoder blocks, and second, changing its initial filter amount, or width. These experiments where performed to identify the optimal architectural configuration for the final surrogate model. 

For the first experiment, the initial filter count was fixed at 16, following the baseline architecture proposed by \citet{hosseiny2021deep}. 
As in the preceding experiments, other hyperparameters were fixed, except for the best configurations identified in those experiments.
Table \ref{tab:depth_metrics} shows the evaluation performance and the complexity of model architectures with depths ranging from 3 to 6. The best performance on the test set was achieved with a depth of 4. For instance, moving from a depth of 3 and an RMSE of 0.0381 m to a depth of 4 reduces the RMSE to 0.0301 m. Increasing the depth further did not yield any additional performance gains. Somewhat as expected, the amount of model parameters drastically increases, for example, from 7.782 million parameters for a depth of 5 to 31.122 million parameters for a depth of 6. Thus, for all subsequent experiments, a depth of 4 was chosen, as it yielded the lowest test RMSE and required a moderate number of parameters.

\begin{table}[htbp]
\centering
\caption{Performance metrics by depth}
\label{tab:depth_metrics}
\begin{tabular}{lcccccc}
\toprule
\textbf{Depth} & \textbf{RMSE (m)} & \textbf{NSE} & \textbf{Parameters (in millions)} \\
\midrule
3 & 0.0381 & 0.9984 &  \textbf{0.483} \\
\textbf{4} & \textbf{0.0301} & \textbf{0.999} & 1.944  \\
5 & 0.0326 & 0.9988 & 7.782  \\
6 & 0.0348 & 0.9987 &  31.122 \\
\bottomrule
\end{tabular}
\end{table}

Following the first half of this experiment, the model-width experiment was conducted. Based on the previous experiment, the model depth was set to 4, as this was found to be optimal. 
Table \ref{tab:width_metrics} shows the evaluation performance and the complexity of model architectures with widths ranging from 8 to 32. A clear trend is observed: decreasing test RMSE with increasing model width. For instance, moving from a width of 16 with a test RMSE of 0.0301 to a width of 32 resulted in an RMSE of 0.0219. Additionally, as expected, the number of parameters increases with a larger width. Despite the significant increase in parameters, the performance gains from the 32-filter model were considered substantial enough to justify its selection for all subsequent experiments. 

\begin{table}[htbp]
\centering
\caption{Performance metrics by width}
\label{tab:width_metrics}
\begin{tabular}{lcccccc}
\toprule
\textbf{Width} & \textbf{RMSE (m)} & \textbf{NSE} & \textbf{Parameters (in millions)} \\
\midrule
8 & 0.0484 & 0.9974 &  \textbf{0.487} \\
16 & 0.0301 & 0.999 &  1.944 \\
32 & \textbf{0.0219} & \textbf{0.9995} & 7.765 \\
\bottomrule
\end{tabular}
\end{table}

Concluding these model architecture experiments, a model architecture with depth 4 and width 32 was chosen as architecture for all subsequent experiments.

\subsubsection{Cross-Validation}
\label{subsec:results_crossval}
\begin{figure*}[h!]
    \centering
    \includegraphics[width=0.7
    \textwidth]{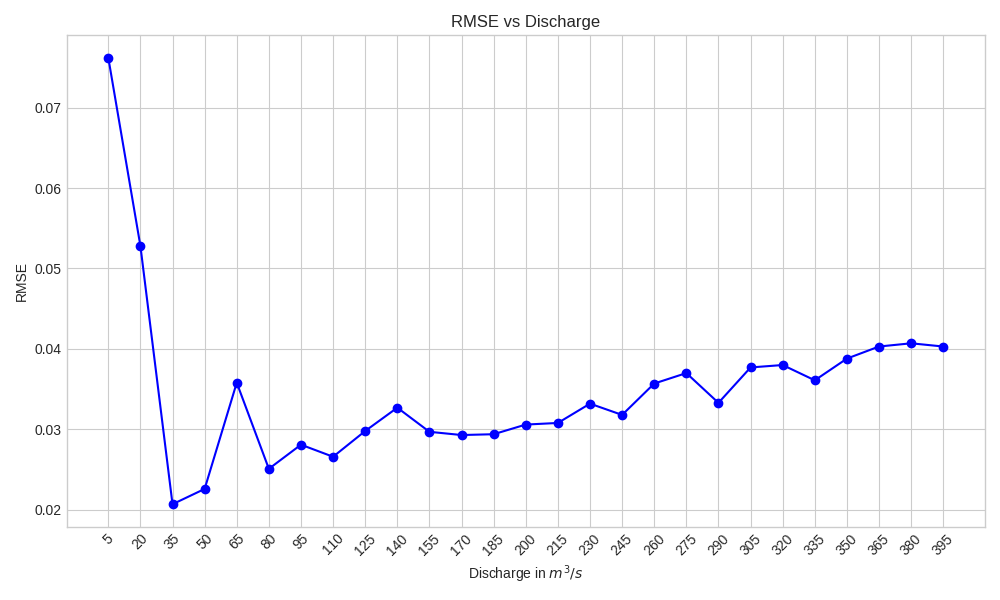}
    \caption{Test RMSE's for the cross-validation experiment.}
    \label{fig:rmse_cross}
\end{figure*}

The following experiment was designed to test for a discharge-related bias in the model and to evaluate its interpolation capabilities across hydrological discharges robustly. Therefore, a leave-one-out cross-validation experiment was conducted. One discharge was selected as the validation set, and the remaining N-1 discharges were used for training. This process was repeated N times for the complete discharges presented in Table \ref{tab:data_splits} for the source domain Beyenburg (BEY). To stay within the computational time limit, each fold was run for 200 epochs. Additionally, the cross-validation experiment was evaluated with the standard performance metrics.
Figure \ref{fig:rmse_cross} shows the test RMSE for each left-out discharge. The results reveal that the model has strong interpolation capabilities across most of the tested values, with the RMSE stabilizing between 0.03 m and 0.04 m. The figure also shows specific weaknesses, with the highest errors occurring at very low discharges (5 $m^3$/s and 25 $m^3$/s) and another notable error spike around the bankfull discharge of 65 $m^3$/s. Overall, this analysis indicates that the U-Net surrogate is a reliable approximator, though its accuracy is reduced at key hydrological thresholds.

\subsubsection{Evaluation of Patch Overlapping during Inference}
The inference method experiment was designed to determine the optimal inference method for generating full-domain predictions. Because the chosen inference method was also used during validation, and the validation loss influenced the training process (via the scheduler), a distinct model was trained for each of the three strategies. Each model
was then evaluated using the standard performance metrics.
The three strategies tested were: the no-overlap method, which used a stride of 128 corresponding to the patch size; the overlap method, which used a stride of 64 and aggregated overlapping predictions via the mean; and the center-crop method, which used only the central 64 $\times$ 64 pixels of each patch for the final prediction. 

The center crop method yielded the lowest RMSE of 0.037 m among the three tested methods. Additionally, this method also yielded the highest NSE score of 0.9985. The overlap method performed slightly worse with an RMSE of 0.0443, followed by the least effective method, the no-overlap, with an RMSE of 0.0487.
Additionally, Table \ref{tab:inference_strategies} shows the average time required for each method to generate a full prediction for an image with size 849 $\times$ 2041. As expected, the center crop method took the longest, with an average of 0.236 sec per image, compared to the fastest method, no-overlap, with an average of 0.067 sec per image. 

\begin{figure*}[htbp] 
    \centering
    \begin{subfigure}{0.7\textwidth}
        \centering
        \includegraphics[width=\linewidth]{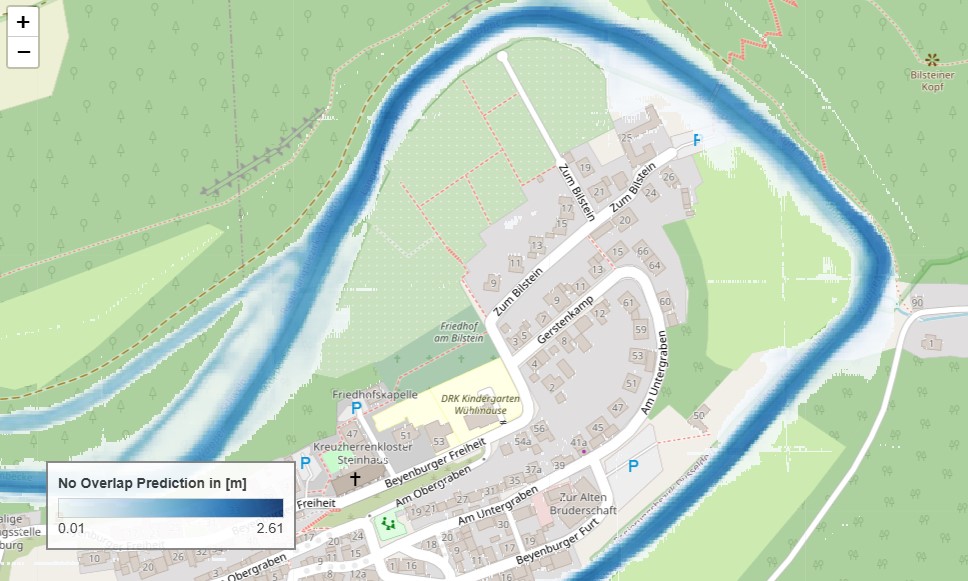}
        \caption{Prediction produced by the no overlap method.}
        \label{fig:no_overlap}
    \end{subfigure}
    \hfill 
    
    \begin{subfigure}{0.7\textwidth}
        \centering
        \includegraphics[width=\linewidth]{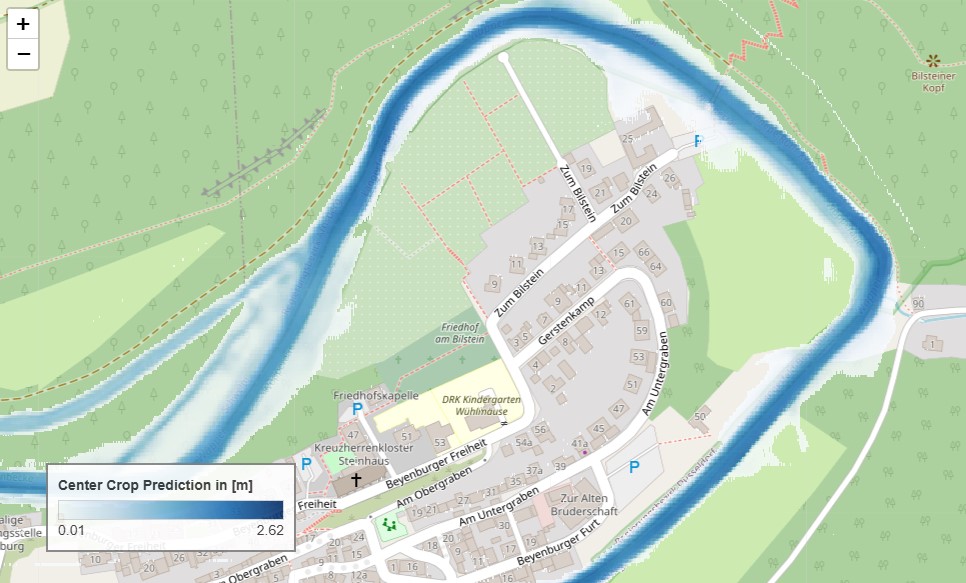}
        \caption{Prediction produced by the center crop method.}
        \label{fig:center_crop}
    \end{subfigure}
    
    \caption[Visual comparison of inference methods]{Visual comparison of inference methods for a discharge of $65 m^3/s$.  Figure (a) shows the surrogate's prediction using the no overlap method in the BEY domain, while Figure (b) shows a prediction produced by the center crop method.}
    \label{fig:inference_comparison}
\end{figure*}

A qualitative comparison of the predictions reveals the root cause for the performance differences observed in Table \ref{tab:inference_strategies}. Fig. \ref{fig:inference_comparison} shows subtle but visible grid-like patterns of white lines in the no overlap prediction (Figure \ref{fig:no_overlap}). These patterns are entirely absent in the center crop prediction (Figure \ref{fig:center_crop}). Since the center crop method achieved the best performance and required only slightly longer inference time than the second-best method, overlap, it was adopted for validation and testing in all subsequent experiments.
\begin{table}[htbp]
\centering
\caption[Performance comparison of inference strategies]{Performance comparison of inference strategies across all locations}
\label{tab:inference_strategies}
\begin{tabular}{llccc}
\toprule
\textbf{Strategy} & \textbf{RMSE (m)} & \textbf{NSE} & \textbf{Inference Time (s)} \\
\midrule
no overlap & 0.0487 & 0.9974 & \textbf{0.067} \\
overlap & 0.0443 & 0.9979 & 0.170 \\
\textbf{center\_crop} & \textbf{0.037} & \textbf{0.9985} & 0.236 \\
\bottomrule
\end{tabular}
\end{table}
The patch inference strategy revealed that while the center-crop strategy does have the highest inference duration of these three methods, it did provide the best results on the primary study area BEY.

\subsubsection{Zero-shot Forecasting Evaluation}

The next step is to evaluate the pre-trained BEY model's generalization capability on BEY under other catchment conditions. To this end, the surrogate model
Table \ref{tab:zero_shot} shows the zero-shot performance of the pre-trained surrogate model. For both unseen locations, the pre-trained surrogate model performed quite poorly, yielding a NSE of -1.5187 for the KLU domain and -1.1539 for the KOL domain. These results indicate that despite the good performance on the original training study area BEY, the model is not generalizing to new locations in a zero-shot scenario.

\begin{table*}[htbp]
\centering
\caption[Performance metrics on primary test domain BEY]{Performance metrics on primary test domain BEY as well as zero-shot performance metrics on unseen domains KLU and KOL}
\label{tab:zero_shot}
\begin{tabular}{lcccccc}
\toprule
\multicolumn{2}{c}{\textbf{BEY}} & \multicolumn{2}{c}{\textbf{KLU}} & \multicolumn{2}{c}{\textbf{KOL}} \\
\cmidrule(lr){1-2} \cmidrule(lr){3-4} \cmidrule(lr){5-6}
\textbf{RMSE (m)} & \textbf{NSE} & \textbf{RMSE (m)} & \textbf{NSE} & \textbf{RMSE (m)} & \textbf{NSE} \\
\midrule
 0.0227 & 0.9994  & 1.2098 & -1.5187 & 1.1455 & -1.1539 \\
\bottomrule
\end{tabular}
\end{table*}

\subsubsection{Qualitative Assessment of the Final Surrogate Model}
\label{subsec:qualitative_results}

\begin{figure*}[h!] 
    \centering
    \begin{subfigure}{0.85\textwidth}
        \centering
        \includegraphics[width=\linewidth]{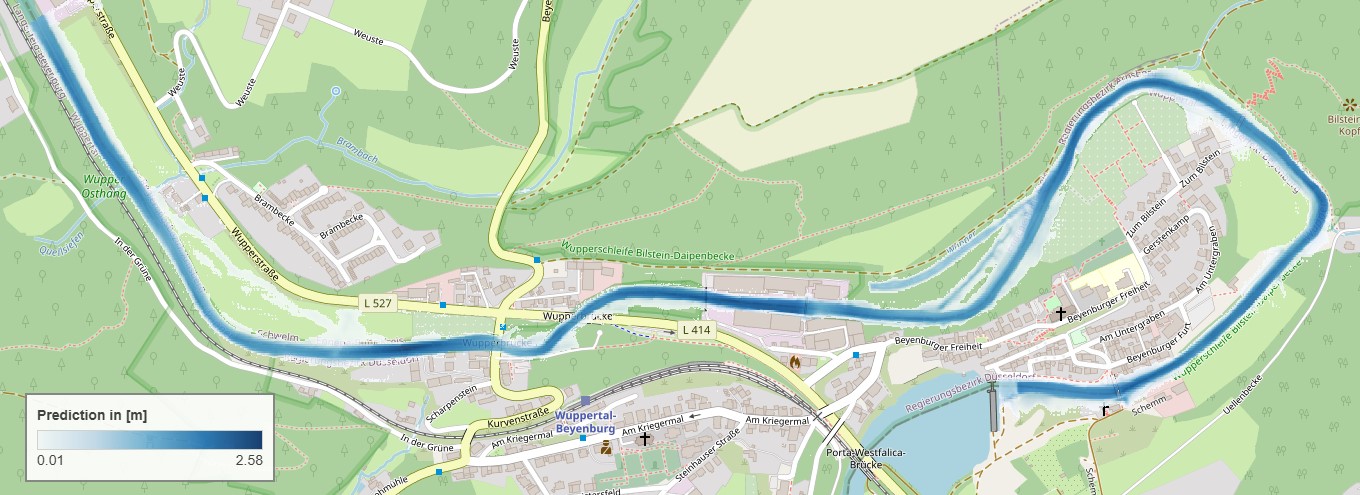}
        \caption{Surrogate Model Prediction}
        \label{fig:prediction_map}
    \end{subfigure}
    \hfill 
    
    \begin{subfigure}{0.85\textwidth}
        \centering
        \includegraphics[width=\linewidth]{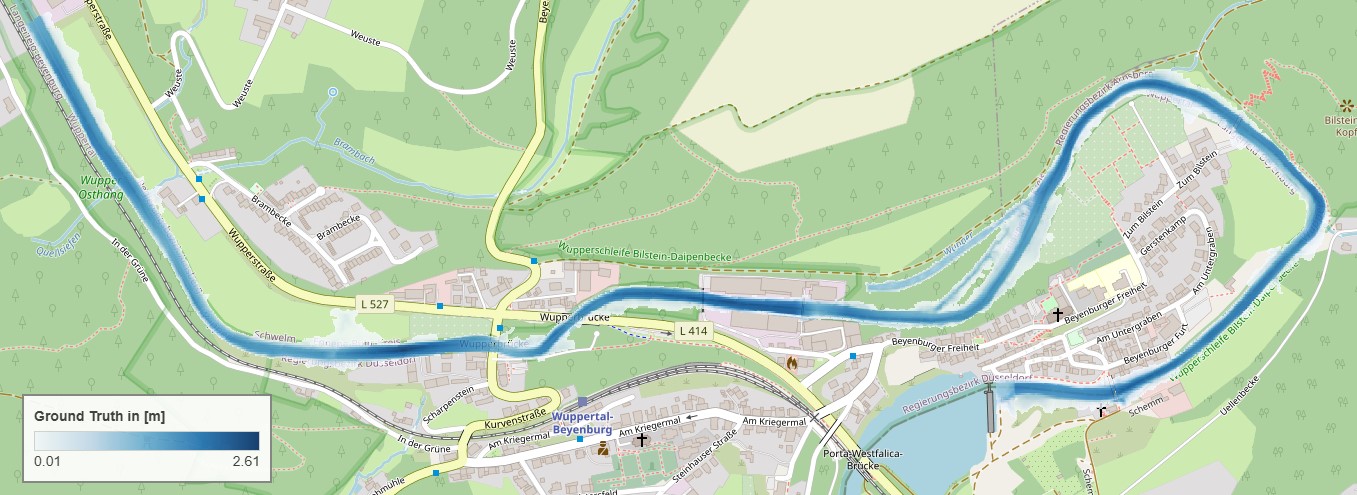}
        \caption{Hydraulic Model (Ground Truth)}
        \label{fig:ground_truth_map}
    \end{subfigure}
    
    \caption[Qualitative performance of the surrogate model]{Qualitative performance of the surrogate model's ability to approximate the hydraulic model. Figure (a) shows the surrogate's prediction for a test discharge of 65 $m^3$/s in the BEY domain, while Figure (b) shows the corresponding ground truth.}
    \label{fig:qualitative_comparison}
\end{figure*}

To complement the quantitative metrics presented in the previous sections, this section provides a qualitative assessment of the final surrogate model's performance. The objective is to diagnose not only the magnitude but also the characteristics and spatial patterns of the prediction errors. 
Direct visualizations of the surrogate model's prediction and the hydraulic model's ground truth for a river discharge of 65 $m^3/s$ are shown in Fig. \ref{fig:prediction_map} and Fig. \ref{fig:ground_truth_map}, respectively. This comparison reveals that the surrogate model successfully captures the overall spatial extent and inundation patterns of the hydraulic model's simulation. While the hydraulic model simulates a maximum water level of 2.61 m, the surrogate model predicts 2.58 m, a slight underestimation relative to the hydraulic model. Errors below 0.01 m were excluded from the following analysis.

\begin{figure*}[h!]
    \centering
    \includegraphics[width=0.7\textwidth]{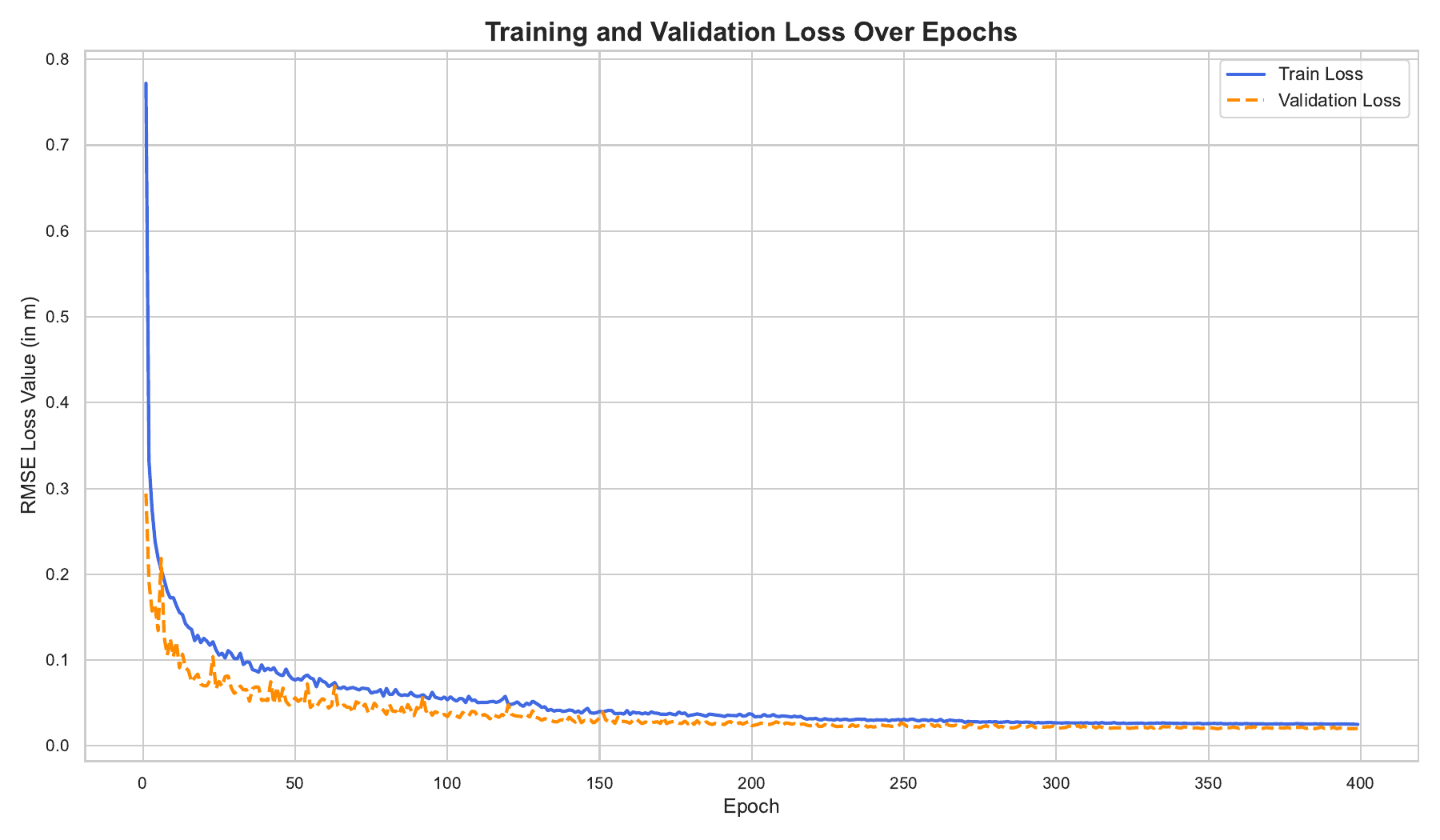}
    \caption[Training and validation loss of surrogate model]{Training and validation loss per epoch of the final surrogate model.}
    \label{fig:surrogate_loss}
\end{figure*}

\begin{figure*}[h!]
    \centering
    \includegraphics[width=0.85\textwidth]{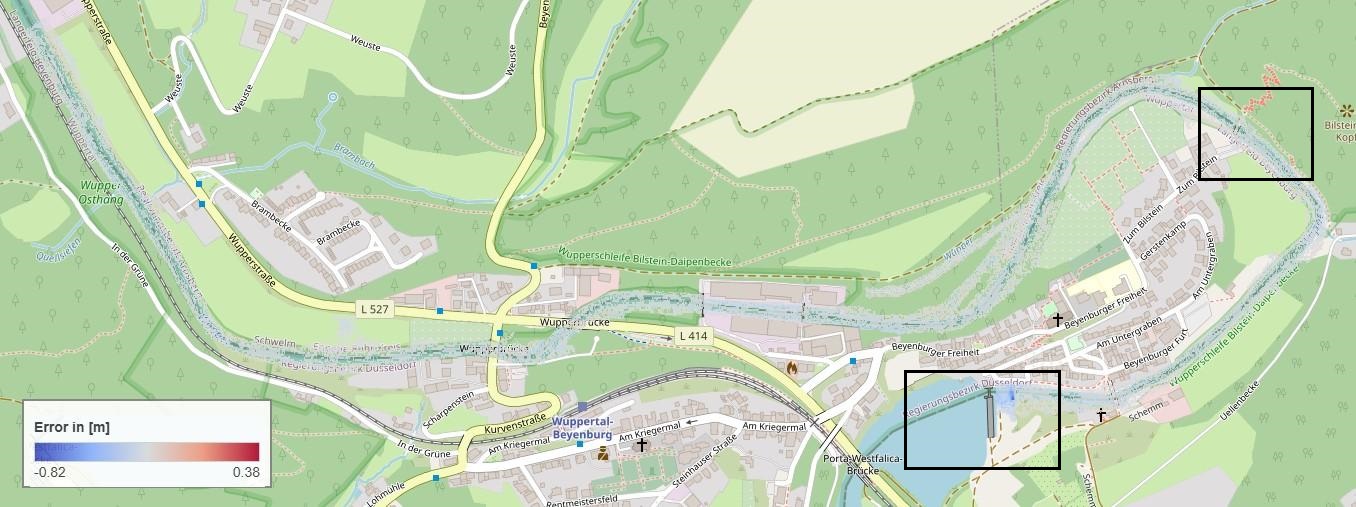}
    \caption[Spatial distribution of prediction error]{Spatial distribution of prediction error for the test discharge of 65 m³/s in the BEY domain. Red areas indicate over-prediction by the surrogate model, while blue areas indicate under-prediction.}
    \label{fig:spatial_error_map}
\end{figure*}

\begin{figure*}[h!]
    \centering
    \includegraphics[width=0.7\textwidth]{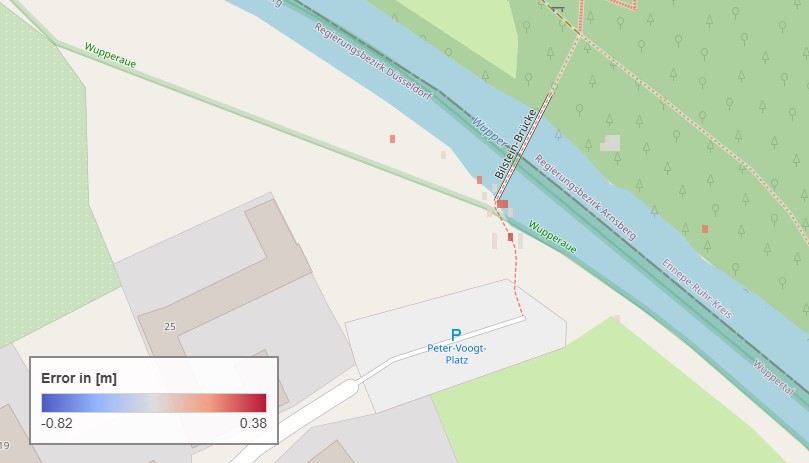}
   \caption[Prediction error near bridge]{Spatial distribution of prediction error near a bridge structure. The figure shows a cluster of large positive errors (red), indicating that the surrogate model significantly over-predicted the water level in this area.}
   \label{fig:bridge}
\end{figure*}

To better understand the specific locations and patterns of these prediction errors, a spatial error map was generated by subtracting the surrogate's prediction from the ground truth. This spatial error map in Fig. \ref{fig:spatial_error_map} confirms the overall good accuracy of the surrogate model. The map reveals that while errors are present, their magnitude is generally low across most of the inundated area. Additionally, most of the larger errors are not distributed throughout the study area but are clustered in specific locations, such as near the dam wall or bridges. 
Another example of an area with larger errors is shown in Fig. \ref{fig:bridge}. Here, the model over-predicts the water level to a large extent. These overpredictions are spatially close to a bridge, indicating that this structure might have influenced the model's performance. 
To finish off the analysis of the final prediction errors,  Fig. \ref{fig:surrogate_loss} illustrates the training process of the final surrogate model. The plot displays the training and validation RMSE over 400 epochs, showing a stable convergence after approximately 250 epochs. 

Additionally, the small and consistent gap between the training and validation loss curves indicates that the model did not suffer from significant overfitting and learned patterns that generalize well to unseen validation data. Interestingly, the validation loss curve is below the training loss curve. Normally, this would not be expected, but in this scenario, the model had more difficulty during training than during validation. This is likely due to the use of random patches during training. However, during validation, the model generated a full-image prediction using the inference method identified in the experiment. 
During validation, the model did not exhibit edge artifacts, unlike during training.

Overall, however, this plot indicates that the surrogate model learned to generalize effectively across the range of discharge values for the BEY domain.
Increasing model width generally improved performance on the BEY test set. However, this performance increase was not observed for depth, where the best performance was achieved at a depth of 4. Therefore, the final model was constructed with a depth of 4 and an initial number of filters of 32. This model provided a good balance between complexity and performance on the test domain BEY.

\subsection{Ablation Study of Patch Parameters and Target Normalization}

An ablation study was conducted to determine the impact of the patch parameters and normalization during training.

\subsubsection{Patch Size}
The following analysis investigated the trade-off between performance and computational complexity when using different patch sizes. 
For this purpose, two models were trained with the same baseline architecture. 
The tested patch sizes were 128 $\times$ 128 and 256 $\times$ 256 pixels. Training a model with $512 \times 512$ patches was not feasible because it exceeded the capacity of a single GPU. The same performance metrics 
were used to evaluate each model. Additionally, the center crop method was used to validate and test the models as determined by the previous experiment. 
Tables \ref{tab:patch_size_metrics} and \ref{tab:patch_size_performance} show performance metrics on the test set and training and inference times for each patch size configuration, respectively. 
\begin{table}[!htbp]
\centering
\caption{Performance metrics by patch size}
\label{tab:patch_size_metrics}
\begin{tabular}{lcccccc}
\toprule
\textbf{Patch Size} & \textbf{RMSE (m)} & \textbf{NSE} \\
\midrule
128 x 128 & 0.037 & 0.9985 \\
\textbf{256 x 256} & \textbf{0.031} & \textbf{0.999}  \\
\bottomrule
\end{tabular}
\end{table}
\begin{table*}[htbp]
\centering
\caption{Computational performance by patch size}
\label{tab:patch_size_performance}
\begin{tabular}{lcccc}
\toprule
\textbf{Patch Size} & \textbf{Training time (min)} & {\textbf{Inference time per image (min)}}\\
\midrule 
\textbf{$128 \times 128$} & \textbf{61.51} & \textbf{0.214} \\
$256 \times 256$ & 183.6 & 0.696  \\
\bottomrule
\end{tabular}
\end{table*}
The 256-model yielded an RMSE of 0.031 m, compared with 0.037 m for the 128-model. Additionally, the 256-model achieved a higher NSE of 0.999. The training time increases from 61.51 min for the 128-model to 183.6 min for the 256-model. A similar increase is observed in the average inference time per image, from 0.214 sec to 0.696 sec.
Given the slight performance improvement of the 256-model despite the longer training and inference times, a patch size of 256 was chosen for all subsequent experiments.
A larger patch size of 256 yielded better performance than a smaller patch size of 128. However, this performance increase came with an almost triple increase in both inference and training time.

\subsubsection{Target normalization}
\label{subsec:results_normalization}
Normalizing the target variable yielded a slight increase in RMSE from 0.0335 to 0.0368 without target normalization. This was established by comparing two models, one trained with normalized targets, the other with targets in the original scale. Both models were compared with the standard performance metrics. 
Following best practices, the normalization statistics were computed on the training targets to prevent information leakage from the test images into the training. 
The statistics were applied to normalize the targets of the training, validation as well as test set through min-max scaling. To provide easy-to-interpret results, targets and predictions were scaled back to compute the performance metrics. 
Despite the slight decrease in RMSE, we chose for target normalization to create a more robust and generalizable model. This acceptable trade-off is made to better prepare the model for the transfer learning experiments.

\begin{table}[htbp]
\centering
\caption{Impact of target normalization}
\label{tab:target_normalization}
\begin{tabular}{lcccccc}
\toprule
\textbf{Method} & \textbf{RMSE} & \textbf{NSE} \\
\midrule
With Normalization & 0.0368 & 0.9985 &  \\
\textbf{Without Normalization} & \textbf{0.0335} & \textbf{0.9988}  \\
\bottomrule
\end{tabular}
\end{table}

While training without target normalization yielded a slightly lower RMSE on the source domain, normalization was adopted to create a more robust, scale-invariant model. 

\subsubsection{Patch Amount}
\label{subsec:results_amount}
To quantify the extent to which the number of sampled patches per training image influences the model's performance, an experiment was designed to test multiple patch-count configurations. For this, multiple models were trained with varying numbers of patches sampled per image, ranging from 100 to 700. These models were trained until convergence following the early stopping logic 
and evaluated using the standard metrics. 
Additionally, the validation loss per update step for the selected patch amount configuration was compared. Comparing the validation loss per update step rather than per epoch was necessary to ensure a fair comparison. A model trained with 100 patches per training image effectively sees less training data per epoch than a model trained with 700 patches. Thus, any difference in model performance when evaluating validation loss per epoch could also be explained by a larger training volume. Since all validation loss curves per model update step are very close, and the visualization resulted in a dense plot, we decided to plot these validation curves for models trained with patch sizes of 200, 400, and 600.

\begin{figure*}[!htbp]
    \centering
    \includegraphics[width=0.7\textwidth]{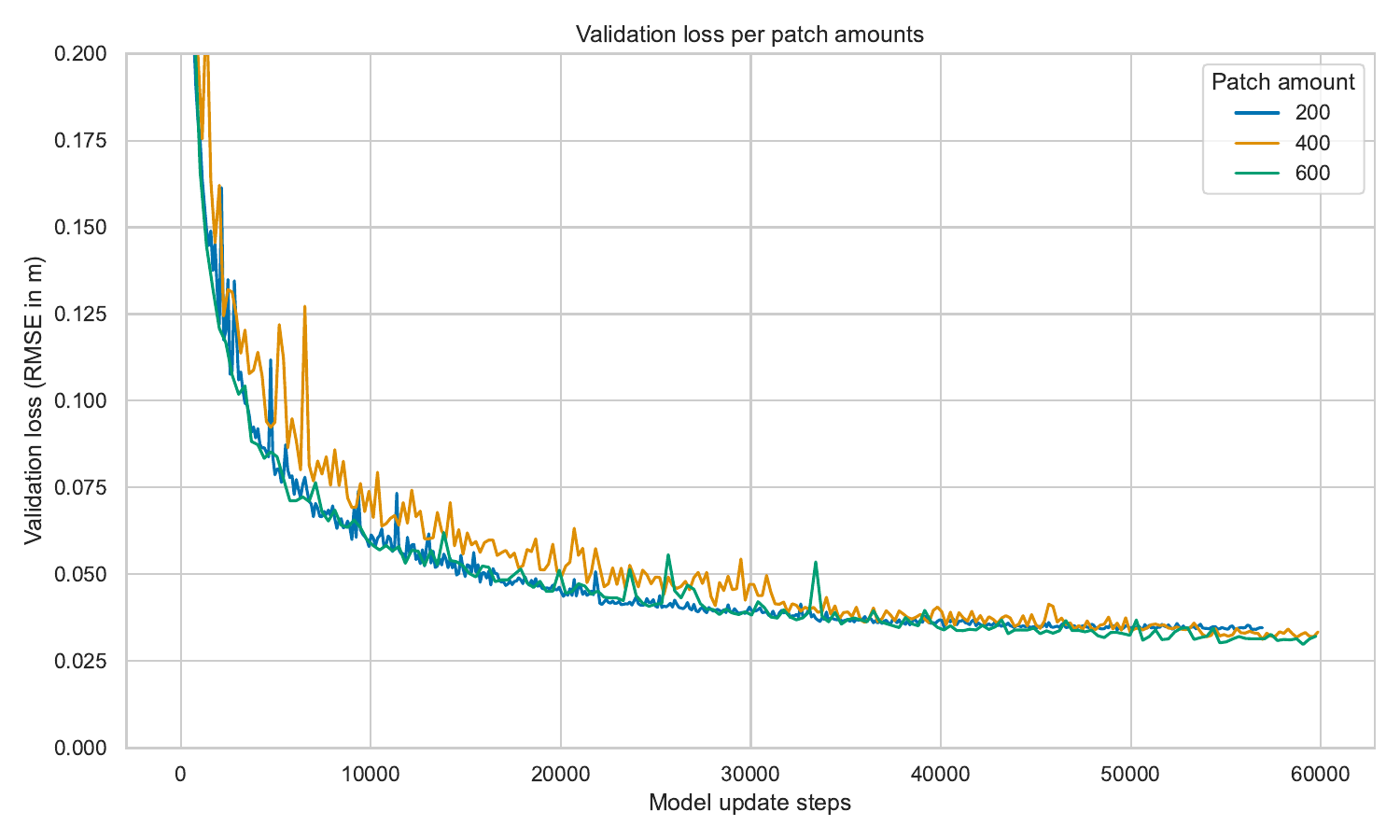}
    \caption[Validation loss per patch amount]{Model performance on validation set for different patch amounts per training image.}
    \label{fig:patch_amount_efficiency}
\end{figure*}

Fig. \ref{fig:patch_amount_efficiency} shows the validation loss per model update step for the aforementioned patch amounts. While the 400-patch model performs slightly worse on the validation set up to 35k update steps, this difference smooths out. For the final model update steps, no clear difference is observed among these three patch amounts. These results show that more patches do not necessarily mean faster learning per step.

\begin{table}[!htbp]
\centering
\caption{Performance metrics by number of patches}
\label{tab:patch_metrics}
\begin{tabular}{lcccccc}
\toprule
\textbf{Number of patches} & \textbf{RMSE (m)} & \textbf{NSE} \\
\midrule
100 & 0.0451 &  0.9978\\
200 & 0.0384 & 0.9984 \\
300 & 0.0334 & 0.9988 \\
400 & 0.0320 & 0.9989 \\
500 & 0.0311 & 0.9989 \\
600 & 0.0268 & 0.9992 \\
700 & 0.0265 & 0.9992 \\
\bottomrule
\end{tabular}
\end{table}

\begin{figure*}[!htbp]
    \centering
    \includegraphics[width=0.7\textwidth]{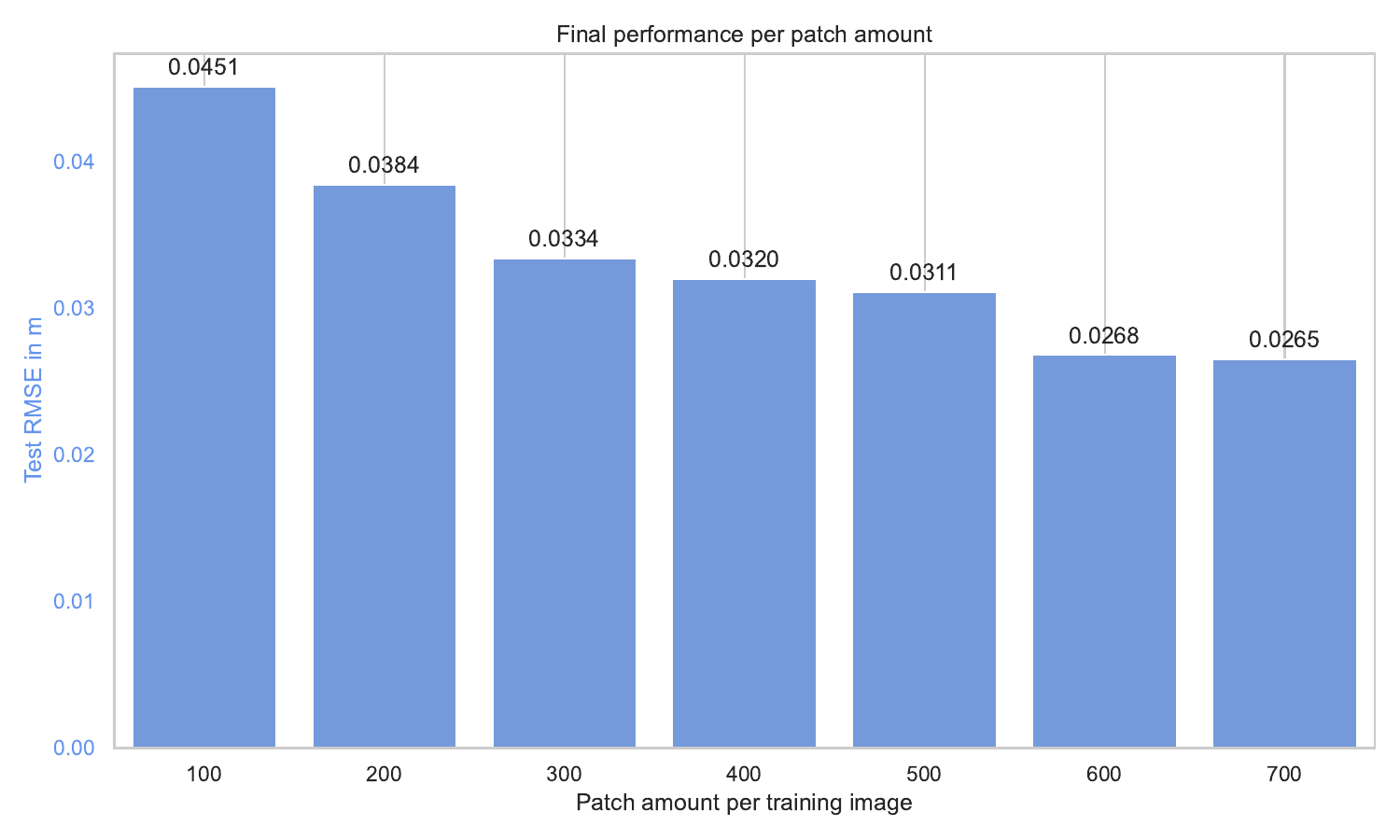}
    \caption{Final performance per patch amount on Beyenburg (BEY) test set.}
    \label{fig:patch_amount_trade_off}
\end{figure*}

Table \ref{tab:patch_metrics} and Fig. \ref{fig:patch_amount_trade_off} show the test RMSE of training with different patch amounts. A trend of decreasing RMSEs with increasing patch amount is shown in Fig. \ref{fig:patch_amount_trade_off}. The patch-100 model is clearly the worst-performing model with a final RMSE of 0.0451 m. The test RMSEs of the remaining models fall within the range 0.0265 m to 0.0384 m, with only slight differences. The trend of decreasing RMSEs shown in the figure must be weighed against the computational cost. As expected, training time increases near-linearly with the number of patches, highlighting the direct cost of sampling more data. As seen in Table \ref{tab:patch_metrics}, increasing the patch number from 600 to 700 yielded only a marginal performance benefit, with the RMSE decreasing from 0.0268 m to 0.0265 m.

Based on these results, a patch amount of 400 was selected. While the 600-patch model offered a significant accuracy gain, the 400-patch model was selected as the optimal trade-off, as it provided strong performance without the considerably higher training cost coming with larger patch amounts.
While using more patches (e.g., 800) yielded further accuracy gains, 400 was selected as the optimal trade-off, providing strong performance on the primary domain BEY without the substantially higher computational cost of the larger patch count.



\section{Discussion}

The model-approximation experiments confirmed that a U-Net architecture can accurately and efficiently approximate a traditional hydraulic model. This final optimized surrogate model successfully approximated the hydraulic model for the BEY study area, achieving high fidelity with a final test RMSE of 0.0227 m and an NSE of 0.994. 
The high accuracy is paired with a significant improvement in computational speed. While the one-time initial training of the surrogate model takes about 5.25 hours, the surrogate generates a full flood map in just 0.696 min for the source domain (BEY) as can be seen in Table \ref{tab:patch_size_performance}. This is a speed-up of over 21 times compared to the 15-minute runtime of the hydraulic model. Furthermore, a key insight from the experiments was the critical role of the inference strategy. Adopting a center-crop method, which mitigates patch-edge artifacts, improved the final RMSE by nearly 25\% over a simple non-overlapping approach.


The superior performance of the center crop methods is directly attributable to their effective mitigation of patch-edge artifacts. These artifacts are unreliable predictions at the patch boundaries and are a fundamental consequence of a CNN's receptive field. Pixels near the edge of a patch have less contextual information than pixels in the center, leading to less reliable predictions at the boundaries. The other two methods perform significantly worse because they mishandle these artifacts. These findings are consistent with recent literature. For instance, the poor performance of the no-overlap method aligns with the work of \citet{guo2021data}, who found that this approach generates many outliers. The no-overlap method includes these artifacts directly in the prediction, whereas the overlap method only slightly dilutes their effect by averaging them. Thus, while the overlap method provides a partial solution by averaging predictions, the center crop performs best because it entirely avoids the problem by discarding unreliable edge predictions, which explains its significantly lower RMSE.

The goal of the patch amount experiment was to investigate the trade-off between model performance and computational cost. For instance, \citet{guo2022data} suggested, that when sampling patches from an image, the amount of sampled pixels should be at least three times larger than the total amount of catchment pixels. Following their heuristic suggests a baseline of approximately 53 patches is needed for the BEY dataset (i.e., $3 \times 1.732.809$ total catchment pixels / $65.536$ pixels per patch $\approx 53$ patches). However, our experiments explored a wider range. The results indicate that a point of diminishing returns is reached around 400 patches per training image. Following Equation \ref{eq:prob_inclusion}, the probability of any given pixel being included in a training patch for a given image is virtually 100\%. $$P_{\text{inclusion}} = 1 - \left(1 - \frac{65.536}{585.513}\right)^{400} \approx 0.9999999999$$ This calculation shows that with a patch amount of 400 per training image, the probability that each pixel will be covered by at least one patch is 1. Further increasing the patch amount offers no meaningful improvement to this already very high probability. Therefore, while the 600-patch model yielded further accuracy gains, the 400-patch model was selected as the optimal trade-off, providing strong performance while avoiding the higher computational costs of training with larger patch amounts. 

\begin{equation}
P_{\text{inclusion}} = 1 - \left(1 - \frac{N}{M}\right)^P
\label{eq:prob_inclusion}
\end{equation}
where \( P_{\text{inclusion}} \) is the probability of a single pixel being included,
\( N \) is the number of pixels in a patch, \( M \) is the total number of valid 
pixels, and \( P \) is the number of patches sampled. 

The model architecture experiments revealed that a deeper model did not improve performance on the flood hazard mapping task. These results were unexpected, as a greater depth is theoretically beneficial for capturing large-scale spatial context, which is particularly relevant in the domain of hydrology \citep{guo2022data,cache2024enhancing}. The reason is that as depth increases, neurons in deeper layers have larger receptive fields, enabling them to capture more contextual information. Additionally, these deeper layers capture high-level information such as the overall shape of the river. However, the experimental results suggest that a depth of 4 was already sufficient to capture the relevant spatial context for the BEY domain. 

The performance degradation of deeper models indicates that the added complexity from depth did not provide additional predictive power and instead made the model more susceptible to overfitting to the single-catchment dataset. In contrast to the depth findings, increasing the network's width significantly improved the model's performance. A plausible explanation for these findings is that the additional capacity of the wider network enabled the model to learn topographical characteristics specific to the BEY study area. This hypothesis is strongly supported by the pre-trained surrogate model's poor zero-shot performance on the two unseen domains KLU and KOL in table \ref{tab:zero_shot}, where it achieved NSE scores of -1.5187 and -1.1539, respectively. These findings align with the work of others, such as \citet{guo2021data}, who proposed that training on a single-catchment dataset can lead to memorization of specific features rather than learning generalizable ones. To address this, they suggest training a model on a diverse, multi-catchment dataset to improve generalization.


The cross-validation experiment revealed that the surrogate model is largely unbiased across all discharges, with the RMSE stabilizing in a low range. However, the analysis also highlights apparent weaknesses at key hydrological thresholds. For instance, a notable error spike at the bankfull discharge (e.g., 65 $m^3$/s in Beyenburg) supports the hypothesis that the model may have overfitted to the more common, almost linear flooding scenarios. This suggests that while the model is effective at interpolation, it struggles to capture the highly non-linear dynamics of critical hydraulic points. 

A visual representation of the surrogate model's prediction confirms this, revealing that the model is quite accurate in capturing the overall extent and depth of flooding, with only a few large errors. As shown in Fig. \ref{fig:bridge} 
these large errors clustered near complex man-made structures such as dams and bridges. For instance, the surrogate model struggled around the area right after the dam, with some large negative errors concentrated there. 
Fig. \ref{fig:bridge} revealed another cluster of errors. These errors with a large positive margin are clustered around a bridge structure. A critical insight gained from this visual error analysis was a likely discrepancy in the preprocessing of the Digital Elevation Model. The DEM used to calibrate the hydraulic model has been thoroughly preprocessed to manually remove bridges and other structures to more accurately guide water flow. However, the DEM used to train the data-driven surrogate model has not been preprocessed, so structures such as this bridge remain present. Future work should prioritize developing a strictly consistent data pipeline between the physics-based model and the surrogate model to reduce errors arising from pre-processing discrepancies, such as inconsistent handling of bridge structures.

\subsection{Limitations}

The primary focus of the first part of the research was to develop an accurate data-driven surrogate model for predicting water levels, similar to that of \citet{kabir2023deep}. For this reason, optimizing input features was considered outside the scope of this work. Throughout this research, the data used for training each surrogate model has been generated by a 2D physics-based model.
This study showed that surrogate models can approximate the outputs of a hydraulic model.  
However, this approach has limitations, including that the model has not been trained or evaluated on field data (e.g., real-world observational flooding data). Similar limitations have been observed by \citet{zahura2020training}, who also suggested that in order to verify the usefulness of such a surrogate model, it is necessary to validate it with real-sensor data (e.g., remote sensing data) \citep{bentivoglio2022deep, MasafuWilliams2024}. Another limitation is the scope of the input data. The surrogate models presented in this work have been trained exclusively on elevation data extracted from a DEM and a constant river discharge. Other essential variables, such as slope or soil saturation, have been neglected as they were outside the scope of this research. A similar limitation has been identified by \citet{guo2022data}. 
In addition to those two variables, \citet{lowe2021u} have identified other important variables that could benefit any data-driven model for flood hazard mapping. In addition, the study's geographical scope was confined to three floodplains within the Bergisches Land. Therefore, these results on generalization may not be applicable to fundamentally different topographies. 
Another limitation is the random patch sampling during training. In this work, patches with at least one valid pixel were deemed sufficient and thus used for training. This approach, however, introduces a form of data imbalance, as a patch with a single valid pixel may exert a much stronger influence on the model's training than a patch filled entirely with valid data. An alternative approach would be, to only sample patches if at least 20\% of the pixels are valid, as is also done by \citet{cache2024enhancing}.

This work  
focused on a U-Net architecture, a specific type of CNN. While the U-Net proved to be effective, this choice excluded other architectures that might have offered different benefits. For instance, increasing the U-Net's depth did not improve performance, suggesting a limitation in its ability to effectively leverage larger spatial context. This finding suggests exploring alternative architectures,
which are specifically designed to capture temporal and large-scale spatial relationships, such as long-short-term memory (LSTM), convolutional-LSTM, and CNN-transformer-based models \citep{xu2023deep, LiaoWang2025, SongGuan2025}.
Although these models yield promising results, they lack physical consistency \citep{LiuWu2026}. To this end, physics-informed neural networks include fluid mechanics into their architectures (e.g., the fluid dynamics are embedded in the loss function) \citep{LiuWu2026, QiAlmeida2024, LiZhao2026}.




\section{Conclusions}\label{section__conclusion}
This paper proposed a deep U-Net framework for flood hazard mapping as a comparable, faster alternative to traditional hydraulic models. For this purpose, this research presented a methodology that included data generation using a patch strategy, a comparison of different inference strategies to assess model robustness, and an ablation study of relevant training parameters.  
The results showed a high-fidelity U-Net surrogate model for the Beyenburg (BEY) study area, which 
achieved a final RMSE of 0.0227 m and a significant computational speed-up of computational time of $21.5$ times compared to the available physics-based model used in this study. 
Furthermore, it laid the groundwork for establishing a real-time flood warning system.
However, a significant limitation of using the surrogate models developed in this study to predict real floods is the lack of performance evaluation on real-world observed flooding data. 

Future research includes robustly assessing the usefulness of a data-driven surrogate model. It is necessary to evaluate it on observed real-field data, such as sensor readings. Additionally, the input data could be enhanced to incorporate topographical data, precipitation data, and discharge hydrographs that simulate the river's actual discharge rather than a constant discharge. These suggestions could yield a more robust model suitable for real-world applications.

\section*{Acknowledgment}
This work is supported by the Ministry of Economic Affairs, Industry, Climate Protection and Energy of the State of North Rhine-Westphalia (MWIKE NRW) and to the KI.NRW (powered by KI.NRW), within the promotion of projects in the innovation and transfer area of North Rhine-Westphalia's leading-edge clusters. The research project is 'HWS 4.0: Bergisches Hochwasserschutzsystem (Bergisch flood protection system)', under the project number KI-HWS-001B.

\bibliographystyle{IEEEtranN}
\bibliography{IEEEabrv,ref_add,ref_orig}

\end{document}